\documentclass{article} 

\usepackage[T1]{fontenc}

\usepackage{geometry}
\usepackage{subcaption}

\usepackage[russian,english]{babel} 
\usepackage{amssymb}
\usepackage{amsmath}
\usepackage{multirow}
\usepackage{txfonts}
\usepackage{multicol}
\usepackage{tabularx}
\usepackage{booktabs,adjustbox }
\usepackage{mathdots}
\usepackage{graphicx} 
\usepackage{fancyhdr}
\usepackage{hyperref}
\usepackage{float}

\usepackage{caption}
\makeatletter
\def\@xfootnote[#1]{%
  \protected@xdef\@thefnmark{#1}%
  \@footnotemark\@footnotetext}
\makeatother

\usepackage{sectsty}
\subsectionfont{\normalfont\itshape}
\subsubsectionfont{\normalfont\itshape}

\begin{document}


\begin{tabular}{p{1.1in}p{4.5in}p{1.2in}}  
\hspace{-1cm}
\noindent
\begin{tabular}{c}  \includegraphics[width=2.9cm]{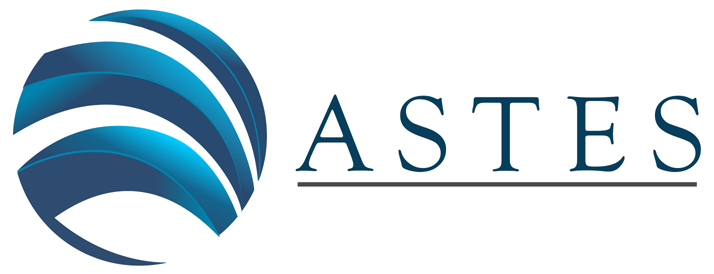}\end{tabular} 	& \vspace{-0.5cm} \centering \textit{Advances in Science, Technology and Engineering Systems Journal \newline Vol. 5, No. 2, XX-YY (2020)} \\   \href{http://www.astesj.com}{www.astesj.com}  
	& \vspace{-0.6cm}  \rule{1.2in}{1.5pt} \vspace{-0.2cm} \newline \centering \textbf{ ASTES Journal \newline ISSN: 2415-6698} \newline \rule{1.2in}{1.7pt} 
\end{tabular}

\vspace{1.8cm}

\noindent \textbf{\LARGE{\setlength\itemsep{0pt}Classification of handwritten names of cities and Handwritten text recognition using various deep learning models}}

\vspace{0.2cm}

\noindent Daniyar Nurseitov${}^{1,2}$\!\!, Kairat Bostanbekov${}^{1,2}$\!\!, Maksat Kanatov${}^{1,2}$\!\!, Anel Alimova ${}^{1,2}$\!\!, Abdelrahman Abdallah\footnote[*]{Abdelrahman Abdallh, Satbayev University, Almaty, Kazakhstan \& abdoelsayed2016@gmail.com }${}^{2,3}$\!\!, Galymzhan Abdimanap ${}^{2,3}$

\vspace{0.2cm}
 \textit{${}^{1}$ Satbayev University, Almaty, Kazakhstan}

\vspace{0.2cm}
 \textit{${}^{2}$National Open Research Laboratory for Information and Space Technologies, Almaty, Kazakhstan}

\vspace{0.2cm}
\textit{${}^{3}$MSc Machine Learning \& Data Science , Satbayev University, Almaty, Kazakhstan}

\vspace{0.3cm}

\begin{tabular}{p{1.7in} p{0.1in} p{5.0in} }
A R T I C L E \hspace{0.1cm} I N F O &  & A B S T R A C T \\ 
 \cline{1-1}  \cline{3-3} \setlength\itemsep{0pt} \vspace{-0.1cm}
\textit{Article history:
	\newline Received:
	\newline Accepted:
	\newline Online:
	\newline \rule{1.78in}{0.5pt} 
	Keywords: 
	\newline Deep Learning
	\newline convolutional neural networks   
	\newline recurrent neural networks
	\newline Russian handwriting recognition
	\newline Connectionist Temporal Classification
	}
 \newline \newline  & & \vspace{-0.1cm} 
\textit{This article discusses the problem of handwriting recognition in Kazakh and Russian languages. This area is poorly studied since in the literature there are almost no works in this direction. We have tried to describe various approaches and achievements of recent years in the development of handwritten recognition models in relation to Cyrillic graphics. The first model uses deep convolutional neural networks (CNNs) for feature extraction and a fully connected multilayer perceptron neural network (MLP) for word classification. The second model, called SimpleHTR, uses CNN and recurrent neural network (RNN) layers to extract information from images. We also proposed the Bluechet and Puchserver models to compare the results. Due to the lack of available open datasets in Russian and Kazakh languages, we carried out work to collect data that included handwritten names of countries and cities from 42 different Cyrillic words, written more than 500 times in different handwriting. We also used a handwritten database of Kazakh and Russian languages (HKR). This is a new database of Cyrillic words (not only countries and cities) for the Russian and Kazakh languages, created by the authors of this work.}\\
\cline{1-1}  \cline{3-3}
\end{tabular}

\vspace{0.5cm}

\begin{multicols}{2}

\section{ Introduction}
This paper is an extension of work originally presented at International Conference on Electronics, Computer and Computation (ICECCO)\cite{daniyar2019classification}

Handwriting text recognition (HTR) is the process of changing handwritten characters or phrases into a format that the computer understands. It has an active network of educational researchers studying it for the past few years as advances in this subject help to automate different types of habitual tactics and office work. An example could be a painstaking seek of a scientific document inside heaps of handwritten ancient manuscripts by a historian, which is requires  a huge amount of time. 

Converting these manuscripts right into a virtual layout using HTR algorithms could permit the historian to find the data within a few seconds. Other examples of ordinary work that need automation will be the tasks associated with signature verification, author recognition, and others. The digitized handwriting text could make contributions to the automation of many corporations’ business approaches, simplifying human work. Our nation postal carrier, as an instance, does not have an automated mail processing gadget that recognizes handwritten addresses on an envelope. The operator has to work manually with the data of any incoming correspondence. Automation of this commercial cycle of mail registration might dramatically decrease postal carrier fees on mail shipping.


The key advances in HTR in mail communication are primarily aimed at finding solutions to the problems of recognition of the region of interest in the images, text segmentation, elimination of interference when working with text background noises, such as missing or ambiguous bits, spots on paper, detection of skew.

The whole cycle of recognizing handwritten addresses of a written correspondence using machine learning from start to end will consist of the following steps:
\begin{itemize}
	\item Letters are put face-up on a conveyor in motion.
	\item A snapshot is taken at a certain place of the conveyor.
	\item The machine handles the snapshot and issues addresses to both the sender and the receiver.
	\item The address is passed to the sorting and tracking system.
\end{itemize}

Any supervised machine learning problem requires labeled input data on which to train the model. In our case, it is necessary to train at least two models: one to determine the areas of the image where the text is located, and the other to recognize words. Forms for collecting handwriting samples by keywords were designed and launched. A set of data was formed from scanned images of the front sides of envelopes with handwritten text for training in determining the areas of interest in the image. A model was trained to detect an area of handwritten text on the face of an envelope. The algorithm for segmentation of the detected text block by lines and words was implemented using the construction of histograms.


Offline handwritten address recognition is a special case of Offline Cursive Word Recognition (CWR). The main difference is that the set of words for recognition is limited to words that can occur in addresses. To solve the problem of handwriting recognition, machine learning methods are used, namely, RNN and CNN in HTR (Bluche\cite{bluche2017gated},Puigcerver\cite{puigcerver2017multidimensional}).

\begin{figure}[H]
	\centering
	\includegraphics[width=\linewidth]{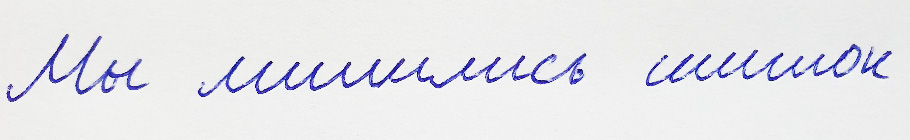}
	\caption{An example of Russian text which difficult to recognize \foreignlanguage{russian}{"Мы лишились шишок”} (“My lishilis shishok”)} 
		\label{fig:russian_texxt}
\end{figure}

Russian and Kazakh languages are very difficult and challenging when it comes in recognizing text language, where writers can write the character contact together like in Figure \ref{fig:russian_texxt}, so the segmentation of characters can be impossible.

This project aims at further study of the challenge of classifying Russian handwritten text and translating handwritten text into digital format. Handwritten text is a very broad concept, and for our purposes we decided to restrict the reach of the project by specifying the meaning of handwritten text.
In this project we took on the task of classification of the handwritten word, which could be like a sequence writing. This research will be combined with algorithms segmenting the word images into a given line image, which in turn can be combined with algorithms segmenting line images into a given image of an entire handwritten page. Our research will take the form of the end-to-end user program and will be a fully functional model that serves the user to solve the problem of transformation of a handwritten text to the digital format. 
The aim of the work is to implement such a recognition handwritten system that will be able to recognize Russian and Kazakh handwriting words written by different writers. We use the HKR database\cite{nurseitov2020hkr} as training, validation and test set. The main contributions covering several key techniques proposed in our system can be highlighted as follows.

\begin{enumerate}
	\item Pre-processing of the snapshot input (noise elimination, horizontal alignment). An unprocessed snapshot is forwarded  as input data at this step. Here the noise is reduced, and  the object’s angle of rotation is measured along the axis perpendicular to the plane.
    \item Segmentation of areas by word within the text. Handwritten words are described at this stage of service, and they are cut into rectangular areas within the text for further recognition.
   \item Recognition of Words. After the snapshot segmentation into separate areas of words has been effective, direct word recognition will begin.
\end{enumerate}

In this article we evaluated models using two methods: in the first method the standard performance measures are used for all results presented: the character error rate (CER) and word error rate (WER)\cite{frinken2014continuous}. The CER is determined as the deviation from Levenshtein, which is the sum of the character substitution (S), insertion (I) and deletions (D) required to turn one string into the other, divided by the total number of characters in the ground truth word (N). Similarly, the WER is calculated as the sum of the number of term substitutions (Sw), insertion (Iw) and deletions (Dw), which are necessary for transformation of one string into the other, divided by the total number of ground-truth terms (Nw). The second method is calculation of the character accuracy rate(CAR) and word accuracy rate(WAR).

This article considers four main models based on artificial neural networks (ANN). Russian and Kazakh handwriting recognition is implemented using Deep CNN\cite{albawi2017understanding}, SimpleHTR model\cite{scheidl2018handwritten}, Bluche\cite{bluche2017gated}, and Puigcerver\cite{puigcerver2017multidimensional}.

The paper is structured as follows: Section 2 describes the related work, Section 3 presents the description of models proposed in the work. Comprehensive results are presented in Section 4, and conclusions in Section 5.



\section{Related Work}
In offline HTR, the input features are extracted and selected from images, then ANN or HMM are used to predict the probabilities and decode them for the final text. The main disadvantage of HMMs is that they cannot predict long sequences of data. 
HMMs have been commonly used for offline HTR because they have achieved good results in automatic speech recognition \cite{el1999hmm,plamondon2000online}. The basic idea is that handwriting can be perceived as a series of ink signals from left to right that is similar to the sequence of acoustic signals in voice. The inspiration for the hybrid HMM research models came from: 
\begin{enumerate}
    \item offline HTR  \cite{bunke2003recognition,bunke2004offline},
    \item offline HTR using conventional HMMs\cite{gorbe2008handwritten},
    \item automatic speech recognition using hybrid HMM/ANN models\cite{bengio1993connectionist,gemello2010hybrid},
    \item and online HTR\cite{graves2008unconstrained}.
\end{enumerate}

On the other hand, RNNs such as gated recurrent unit (GRU)\cite{chung2014empirical} and long short term memory (LSTM)\cite{doi:10.1162/neco.1997.9.8.1735} can solve this problem. RNN models have shown remarkable abilities in sequence-to-sequence learning tasks such as speech recognition\cite{hannun2014deep}, machine tranlation\cite{sutskever2014sequence}, video summarization\cite{srivastava2015unsupervised}, automated question answer\cite{abdallah2020automated} and others.

To transform a two-dimensional image for offline HTR it is necessary to take the image as a vector and forward it to an encoder and decoder. The task is solved by HTR, GRU and LSTM, which take information and feature from many directions. These handwriting sequences are fed into RNN networks. Due to the use of Connectionist Temporal Classification (CTC)\cite{graves2006connectionist} models, the input feature requires no segmentation. One of the key benefits of the CTC algorithm is that it does not need any segmented labeled data. The CTC algorithm allows us to use data alignment with the output.

A new system called Multilingual Text Recognition Networks (MuLTReNets) was proposed by Zhuo\cite{chen2020multrenets}. In particular, the main modules in the MuLTReNets are: the function extractor, script identifier and handwriting recognition. In order to convert the text images into features shared by the document marker and recognizer, the Extractor function method combines spatial and temporal information. The handwriting recognizer adopts LSTM and CTC to perform sequence decoding. The accuracy of script recognition achieves 99.9\%.


A new attention-based fully gated convolutional RNN was proposed by Abdallah\cite{abdallah2020attention}, this model was trained and tested on the HKR dataset\cite{nurseitov2020hkr}. This work shows the effect of the attention mechanism and the gated layer on selecting relative features. Attention is one of the most influential ideas in the Deep Learning community.  It can be used   in image caption\cite{huang2019attention}, automated question answer\cite{abdallah2020automated}, and other problems in deep learning achieving good results. Attention mechanism was first used in the context of Neural Machine Translation using Seq2Seq Models. It also achieved the state-of-the-art for offline Kazakh and Russian handwriting dataset\cite{nurseitov2020hkr}. Atten-CNN-BGRU architecture achieves 4.5\% CER in the first dataset and 19.2\% WER in HKR dataset and 6.4\% CER and 24.0\% WER in the second test dataset.


A multi-task learning scheme in Tassopoulou\cite{tassopoulouenhancing} research, teaches the model to perform decompositions of the target sequence with target units of varying granularity, from fine to coarse. They consider this approach as a way of using n-gram knowledge in the training cycle, indirectly, while the final recognition is made using only the output of the unigram. Unigram decoding of such a multi-task method demonstrates the capacity of the trained internal representations, placed on the training phase by various n-grams. In this research, pick n-grams as target units and play with granularities of the subword level from unigrams till fourgrams. The proposed model, even evaluated only on the unigram task, outperforms its counterpart single-task by absolute 2.52\% WER and 1.02\% CER, in a greedy decoding, without any overhead computing during inference, indicating that an implicit language model is successfully implemented. 

The Weldegebriel\cite{weldegebriel2019new} for classification proposes a hybrid model of two super classifiers: the CNN and the Extreme Gradient Boosting (XGBoost).  CNN serves as  an automatic training feature extractor for raw images for this integrated model, and XGBoost uses the extracted features as an input for recognition and classification. The hybrid model and CNN output error rates are compared to the fully connected layer. In  the classification of handwritten test dataset images, 46.30\% and 16.12\% error rates were achieved, respectively. As a classifier, the XGBoost gave better results than the conventional fully connected layer.

A fully convolutional handwriting model suggested by Petroski\cite{such2018fully} utilizes an unknown length handwriting sample and generates an arbitrary symbol stream. Both local and global contexts are used by the dual-stream architecture and need strong pre-processing steps such as symbol alignment correction as well as complex post-processing steps such as link-time classification, dictionary matching, or language models. The model is agnostic to Latin-related languages using over 100 unique symbols and is shown to be very competitive with state-of-the-art dictionary methods based on standard datasets from IAM\cite{marti2002iam} and RIMES\cite{Grosicki2006}. On IAM, the fine-tuned model achieves 8.71\% WER and 4.43\% CER and hits 2.22\% CER and 5.68\% WER on RIMES.

Also, a fully convoluted network architecture proposed by Ptucha\cite{ptucha2019intelligent} that outputs random length symbol streams from the handwritten text. The stage of pre-processing normalizes the input blocks to a canonical representation that negates the need of expensive recurrent symbol alignment. To correct the oblique word fragments a lexicon is developed and a probabilistic character error rate is added. On both lexicon-based and arbitrary handwriting recognition benchmarks, their multi-state convolutional approach is the first to show state-of-the-art performance. The final convolutional method achieves 8.22\% WER and 4.70\% CER.

Arabic handwritten character recognition based on deep neural network uses CNN models with regularization parameters such as batch normalization to prevent overfitting proposed by Younis\cite{younis2017arabic}. Deep CNN was applied for the AIA9k\cite{torki2014window} and AHCD\cite{el2017arabic} databases, and the accuracy of classification for the two datasets was 94.8\% and 97.6\%, respectively.


\section{Proposed work}
In this section four architectures for Handwriting recognition are discussed:
\begin{enumerate}
    \item Deep CNN models\cite{albawi2017understanding}.
    \item SimpleHTR model\cite{scheidl2018handwritten}.
    \item Bluche\cite{bluche2017gated}.
    \item Puigcerver\cite{puigcerver2017multidimensional}.
\end{enumerate}
\subsection{Data}
\begin{figure}[H]
\begin{subfigure}{0.2\textwidth}
\fbox{\includegraphics[width=0.8\linewidth,height=0.2\linewidth]{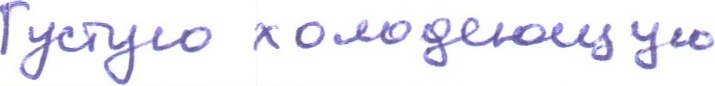}}
\end{subfigure}
\hspace*{\fill}
\begin{subfigure}{0.2\textwidth}
\fbox{\includegraphics[width=0.8\linewidth,height=0.2\linewidth]{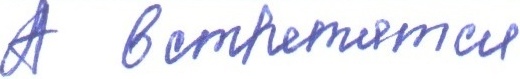}}
\end{subfigure}
\begin{subfigure}{0.2\textwidth}
\fbox{\includegraphics[width=0.8\linewidth,height=0.2\linewidth]{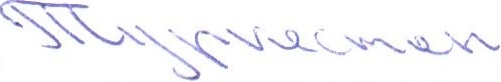}}
\end{subfigure}
\begin{subfigure}{0.2\textwidth}
\fbox{\includegraphics[width=0.8\linewidth,height=0.2\linewidth]{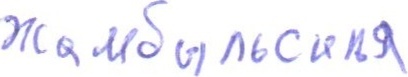}}
\end{subfigure}
\begin{subfigure}{0.2\textwidth}
\fbox{\includegraphics[width=0.8\linewidth,height=0.2\linewidth]{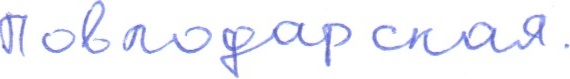}}
\end{subfigure}
\begin{subfigure}{0.2\textwidth}
\fbox{\includegraphics[width=0.8\linewidth,height=0.2\linewidth]{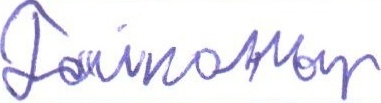}}
\end{subfigure}
\begin{subfigure}{0.2\textwidth}
\fbox{\includegraphics[width=0.8\linewidth,height=0.2\linewidth]{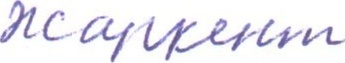}}
\end{subfigure}
\caption{Some Sample of Dataset} 
\label{fig:example}
\end{figure}

In this section, we will describe two types of datasets:
\paragraph{The first}
dataset contains handwritten cites in Cyrillic words. It contains 21,000 images from various handwriting samples(names of countries and cities). We increased this dataset for training by collecting 207,438 images from available forms or samples. 
\paragraph{The second}
HKR for Handwritten Kazakh \& Russian Database\cite{nurseitov2020hkr} consisted of distinct words (or short phrases) written in Russian and Kazakh languages (about 95\% of Russian and 5\% of Kazakh words/sentences, respectively). Note that both languages are Cyrillic written and share the same 33 characters. Besides these characters, there are 9 additional specific characters in the Kazakh alphabet. Some examples of HKR dataset are shown in Figure \ref{fig:example}.

This final dataset was then divided into Training (70\%), Validation (15\%), and Test (15\%) datasets. The test dataset itself was split into two sub-datasets (7.5\% each): the first dataset was named TEST1 and it consisted of words that were not included in the Training and Validation datasets; the other sub-dataset was named TEST2 and consisted of words that were included in the Training dataset but had completely different handwriting styles. The main purpose of splitting the Test dataset into TEST1 and TEST2 datasets was to check the difference in accuracy between recognition of unseen words and words seen in the training stage but with unseen handwriting styles.
\subsection{Deep CNN Models}
\begin{figure}[H]
	\centering
    \includegraphics[width=0.5\textwidth,keepaspectratio]{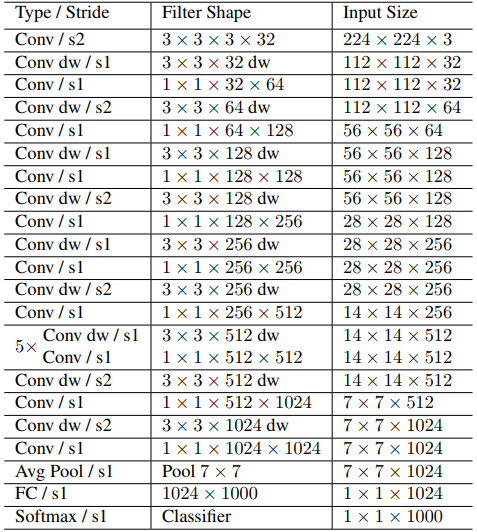}
    \caption{The MobileNet architecture }\label{fig:MobileNet_arch}
\end{figure}

In this experiment, the old approach for classifying images using various deep CNN  models was used. To obtain right distribution of records, pre-processing image strategies and data enlargement methods have been used. Three types of models were considered in the experiment:
The experiment consists of three options:
\begin{enumerate}
	\item Simple model of the CNN\cite{kim2014convolutional},
	\item MobileNet\cite{howard2017mobilenets},
	\item MobileNet with small settings.
\end{enumerate}
\paragraph{Experiment 1}
In this experiment, CNN model was used to train and evaluate the dataset, this model includes 2 conventional layers and a softmax layerr\cite{memisevic2010gated}, which produces probabilities for classification. This model is usually used in character classification like MNIST handwritten digit database\cite{lecun-mnisthandwrittendigit-2010}. The image data feed-forward to the model were (512x61x1). 10\% of the dataset was used for evaluate the model.
\paragraph{Experiment 2 and 3}
In this section we will describe the MobileNet architecture, which consists of 30 layers. The MobileNet model is shown in Figure  \ref{fig:MobileNet_arch}. For more information, see\cite{howard2017mobilenets}. This architecture contains 1x1 convolution layer, Batch Normalization, ReLU activation function, average pooling layer and softmax layer, which is used for classification. 
In Experiment 2 we trained the model for 150 iterations, with Adadelta optimization method \cite{zeiler2012adadelta} where the initial learning step with learning rate (LR) = 1.0 was used.

\subsection{SimpleHTR model}
\begin{figure}[H]
	\centering
	\includegraphics[width=\linewidth]{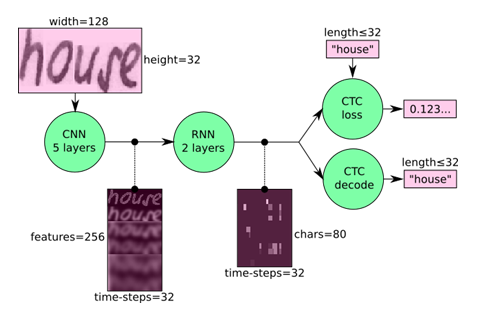}
	\caption{SimpleHTR model, where green icons are operations and pink icons are data streams} 
		\label{fig:SimpleHTR}
\end{figure}

This experiment in our studies used the SimpleHTR system developed by Harald\cite{scheidl2018handwritten}. The proposed system makes use of an ANN, wherein numerous layers of the CNN are used to extract features from the input photo. Then the output of these layers is feed to RNN. RNN disseminates information through a sequence. The RNN output contains probabilities for each symbol in the sequence. To predict the final text, the decoding algorithms are implemented into the RNN output. CTC functions (Figure \ref{fig:SimpleHTR}) are responsible for decoding probabilities into the final text. To improve recognition accuracy, decoding can also use a language model\cite{graves2006connectionist}.
The CTC is used to gain knowledge; the RNN output is a matrix containing the symbol probabilities for each time step. The CTC decoding algorithm converts those symbolic probabilities into the final text. Then, to improve the accuracy,  an algorithm is used that proceeds a word search in the dictionary. However, the time it takes to look for phrases depends on the dimensions of the dictionary, and it cannot decode arbitrary character strings, including numbers.
\paragraph{Operations}

CNN: the input images are fed to the CNN layers. These layers are responsible for extracting features. There are 5x5 filters in the first and second layers and 3x3 filters in the last three layers. They also contain non-linear RELU function and max pooling layer that summarizes images and makes them smaller than the input. Although the height of the image is reduced 2-fold in each layer, the feature maps (channels) are added so as to get the output feature map (or sequence) 32 to 256 in size.

RNN: the feature sequence contains 256 signs or symptoms per time step. The relevant information is disseminated by RNN via these series. LSTM is one of the famous RNN algorithms that carries information at long distances and offers more efficient training functionality than typical RNNs. The output RNN sequence is mapped to a 32x80 matrix.

CTC: receives the output RNN matrix and the predicted text during the neural network learning process, and determines the loss value. CTC receives only the matrix after processing and decodes it into the final text. There should be no more than 32 characters for the length of the main text and the known text.

\paragraph{Data}
Input: This is a size 128 to 32 gray file. Images in the dataset usually do not exactly have this size, so their original dimension is changed (without distortion) until they are 128 in width and 32 in height. The image is then copied to a target image of 128 to 32 in white. Then the gray values are standardized, which simplifies the neural network process.

\subsection{Bluche model}
Bluche model\cite{bluche2017gated} proposes a new neural network structure for modern handwriting recognition as an opportunity to RNNs in multidimensional LSTM. The model is totally based on a deep convolutional input image encoder and a bi-directional LSTM decoder predicting sequences of characters. Its goal is to generate standard, multi-lingual and reusable tasks in this paradigm using the convolutional encoder to leverage more records for transfer learning.
 
The encoder in the Bluche model contains 3x3 Conv layer with 8 features, 2x4 Conv layer with 16 features, a 3x3 gated Conv layer, 3x3 Conv layer with 32 features, 3x3 gated Conv layer, 2x4 Conv layer with 64 features and 3x3 Conv layer with 128 features. The decoder contains 2 bidirectional LSTM layers of 128 units and 128 dense layer between the LSTM layers. Figure 5 shows the Bluche architecture.

\subsection{Puigcerver model}
Modern approaches of  Puigcerver model\cite{puigcerver2017multidimensional} to offline HTR dramatically depend on multidimensional LSTM networks. Puigcerver model has a high level of recognition rate and a large number of parameters (around 9.6 million). This implies that multidimensional LSTM dependencies, theoretically modelled by multidimensional recurrent layers, might not be sufficient, at least in the lower layers of the system, to achieve high recognition accuracy.


The Puigcerver model has three important parts :
\begin{itemize}
\item Convolutional blocks: they include 2-D Conv layer with 3x3 kernal size and 1 horizontal and vertical stride.  number of filters is equal to 16n at the n-th layer of Conv.
\item  Recurrent blocks: Bidirectional 1D-LSTM layers form recurrent blocks, that transfer the input image column-wise from left to right and from right to left. The output of the two directions is concatenated depth-wise.
\item Linear layer: the output of recurrent 1D-LSTM blocks are fed to linear layer to predict the output label. Dropout is implemented before the Linear layer to prevent overfitting (also with probability 0.5).
\end{itemize} 

\begin{figure}[H]
		\centering
    \includegraphics[width=0.4\textwidth,keepaspectratio,height=0.9\textwidth]{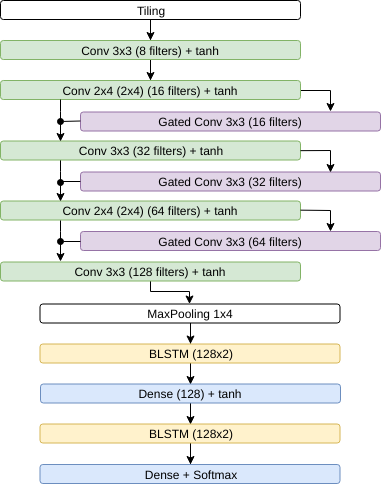}
   \caption{Bluche HTR model}
    \label{fig:Bluchemodels}
\end{figure}
\begin{figure}[H]
		\centering
    \includegraphics[width=0.4\textwidth,keepaspectratio,height=0.5\textwidth]{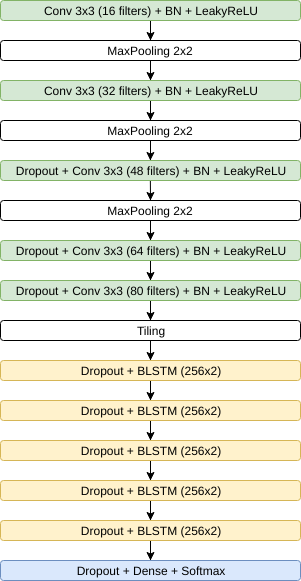}
   \caption{Puigcerver HTR model}
    \label{fig:Puigcervermodels}
\end{figure}

In the new Puigcerver version multidimensional LSTM is used. The difference between regular LSTM networks and multidimensional LSTM is that the former introduce a recurrence alongside the axis of 1-dimensional sequences (as an instance, speaking time-axis or picture writing-direction). In comparison, the latter has a recurrence alongside two axes (generally the x-axis and y-axis in pics). This allows us to use in the latter model unconstrained, two-dimensional information, doubtlessly capturing long-time term dependencies throughout each axis. Figure \ref{fig:Puigcervermodels} shows the Puigcerver architecture.

\subsection{Experiment Materials}
All models have been implemented using the Python and deep learning library called Tensorflow\cite{abadi2016tensorflow}. Tensorflow allows for transparent use of highly optimized mathematical operations on GPUs through Python. A computational graph is defined in the Python script to define all operations that are necessary for the specific computations.

The plots for the report were generated using the matplotlib library for Python, and the illustrations have been created using Inkscape, which is a vector graphics software similar to Adobe Photoshop.
The experiments were run on a machine with 2x “Intel(R) Xeon(R) E-5-2680” CPUs, 4x ”NVIDIA Tesla k20x” and 100 GB RAM. The use of a GPU reduced the training time of the models by approximately a factor of 3, however, this speed-up was not closely monitored throughout the project, hence it could have varied.

\section{Results}
 
This section will discuss the results of the four models on two different datasets. Deep CNN models were trained and evaluated on the first dataset, SimpleHTR model was trained and evaluated on the two datasets and finally Bluche and Puigcerver were trained and evaluated on the second dataset.

\subsection{Deep CNN Result}
For the current experiments, only ten classes from the first dataset were selected: Kazakhstan, Belarus, Kyrgyzstan, Tajikistan, Uzbekistan, Nur-Sultan, Almaty, Aktau, Aktobe, and Atyrau.

\begin{figure}[H]
	\centering
	\begin{subfigure}[b]{0.64\linewidth}
    \includegraphics[width=\textwidth,keepaspectratio]{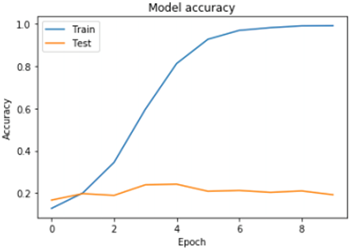}
    \caption{ }
    \end{subfigure}
	\centering
	\begin{subfigure}[b]{0.64\linewidth}
    \includegraphics[width=\textwidth,keepaspectratio]{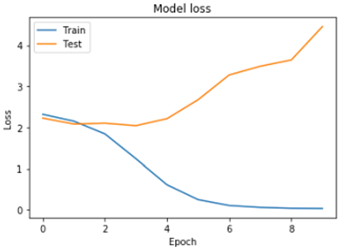}
   \caption{ }
    \end{subfigure}
    
    \caption{First experiment results: (a) Model accuracy, (b) Model error }\label{fig:Experiment_1}
\end{figure}

\paragraph{Experiment 1}
The simple CNN model result is presented. There were 150 iterations in the learning process, and we showed the effects of 10 iterations in the following (Figure \ref{fig:Experiment_1}): In Figure \ref{fig:Experiment_1}, the model goes into a re-learning state after the first iteration, where the results on the training data improve rapidly and the results on the test data, on the contrary, degrade. This means that the model achieves overfitting on this dataset. For this experiment the minibatch of 32 size and the value of the learning rate (lr = 0.01) was used.

\begin{figure}[H]
	\centering
	\begin{subfigure}[b]{0.64\linewidth}
    \includegraphics[width=\textwidth,keepaspectratio]{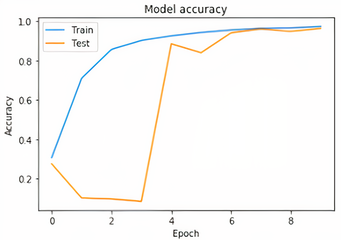}
    \caption{ }
    \end{subfigure}
	\centering
	\begin{subfigure}[b]{0.64\linewidth}
    \includegraphics[width=\textwidth,keepaspectratio]{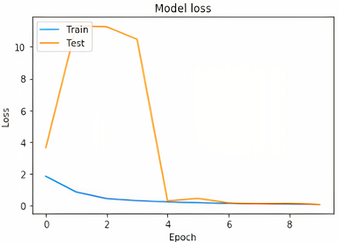}
   \caption{ }
    \end{subfigure}
    \caption{Second experiment results using MobileNet (lr = 1.0): (a) Model accuracy, (b) Model error}\label{fig:Experiment_2}
\end{figure}

\paragraph{Experiment 2 and 3}
Figure \ref{fig:Experiment_2} shows the results after the 10th iteration of learning, and as we can see, the model only learns correctly after the 3rd iteration.

\begin{figure}[H]
	\centering
	\begin{subfigure}[b]{0.64\linewidth}
    \includegraphics[width=\textwidth,keepaspectratio]{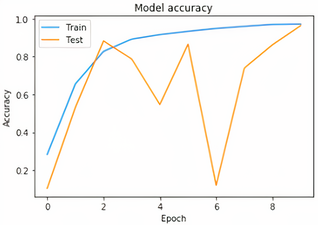}
    \caption{ }
    \end{subfigure}
	\centering
	\begin{subfigure}[b]{0.64\linewidth}
    \includegraphics[width=\textwidth,keepaspectratio]{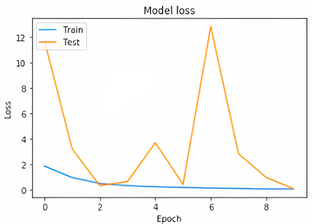}
   \caption{ }
    \end{subfigure}
    \caption{Third experiment results using MobileNet (lr = 0.01): (a) Model accuracy, (b)Model error}\label{fig:Experiment_3}
\end{figure}

The Adadelta optimization method is used and we track the learning behavior, we found that the initial value (lr=1.0) is very large and is not suitable and is to be reduced. Therefore, Experiment 3 was carried out exactly as Experiment 2, the difference is lr=0.01 with minibatch of size 32.The results of Experiment 3 are shown in Figure \ref{fig:Experiment_3}. 

The Adadelta optimization method is used and we track the learning behavior, we found the initial value (lr=1.0) is very large and is not suitable and need to reduce it.Therefore, Experiment 3 was carried out exactly as Experiment 2, the difference is lr=0.01 with minibatch of size 32.The results of Experiment 3 are shown in Figure \ref{fig:Experiment_3}. From Figure \ref{fig:Experiment_3},we can see that the initial iterations have become more correct. However, in the 6th iteration, the model shows a large resonance relative to the entire graph. The reasons for this behavior are still being studied; one of the probable reasons could be small amount of data

The results of the experiments show that the MobileNet Model is better than the simple CNN network because MobileNet is trained on a large dataset for feature extract. This means that the CNN model did not have enough data for training for the feature extract. In order to test this hypothesis, the methods of affinity transformation such as stretching image and other distortions were designed to further increase the data and repeat experiments. After 10 epochs, the training method had the early stop because of overfitting of the models due to the lack of the dataset used in these experiments.


\subsection{SimpleHTR model}
SimpleHTR model is training, validation, and test on two different datasets. In order to launch the model learning process on our own data, the following steps were taken:
\begin{itemize}
\item Words dictionary of annotation files has been created 
\item DataLoader file for reading and pre-possessing the image dataset and reading the annotation file belongs to the images
\item The dataset was divided into two subsets: 90\% for training and 10\% for validation of the trained model.

\end{itemize}
To improve the accuracy and decrease the error rate we suggest the following steps: firstly, increase the dataset by using data augmentation; secondly, add more CNN layers and increase the input size; thirdly, remove the noise in the image and cursive writing style; fourthly, replace LSTM by bidirectional GRU and finally,  use decoder token passing or word beam search decoding to constrain the output to dictionary words.

\paragraph{First Dataset}
For learning on the collected data, the SimpleHTR model was processed, in which there are 42 names of countries and cities with different handwriting patterns. Such data has been increased by 10 times. Two tests were performed: with cursive word alignment and without alignment. After learning the values on data validation presented in Table \ref{tab:veris} were obtained.

This table shows SimpleHTR recognition accuracy for different Decoding Methods (bestpath,beamsearch,wordbeamsearch) 
The best path decoding only uses the NN output and calculates an estimate by taking the most probable character at each position. The beam search only uses the NN output as well, but it uses more data from it and hence provides a more detailed result. The beam search with character-LM also scores character-sequences that further boost the outcome.


\begin{figure}[H]
	\centering
	\includegraphics[width=\linewidth]{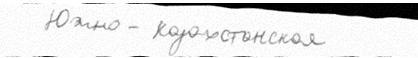}
	\caption{Example of image with phrase "South-Kazakhstan" in Russian} 
		\label{fig:South-Kazakhstan}
	\end{figure}

\begin{figure}[H]
	\centering
	\includegraphics[width=\linewidth]{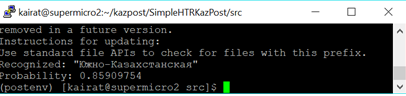}
	\caption{Result of recognition} 
		\label{fig:SimpleHTR_output}
\end{figure}

\begin{figure}[H]
	\centering
	\includegraphics[width=\linewidth]{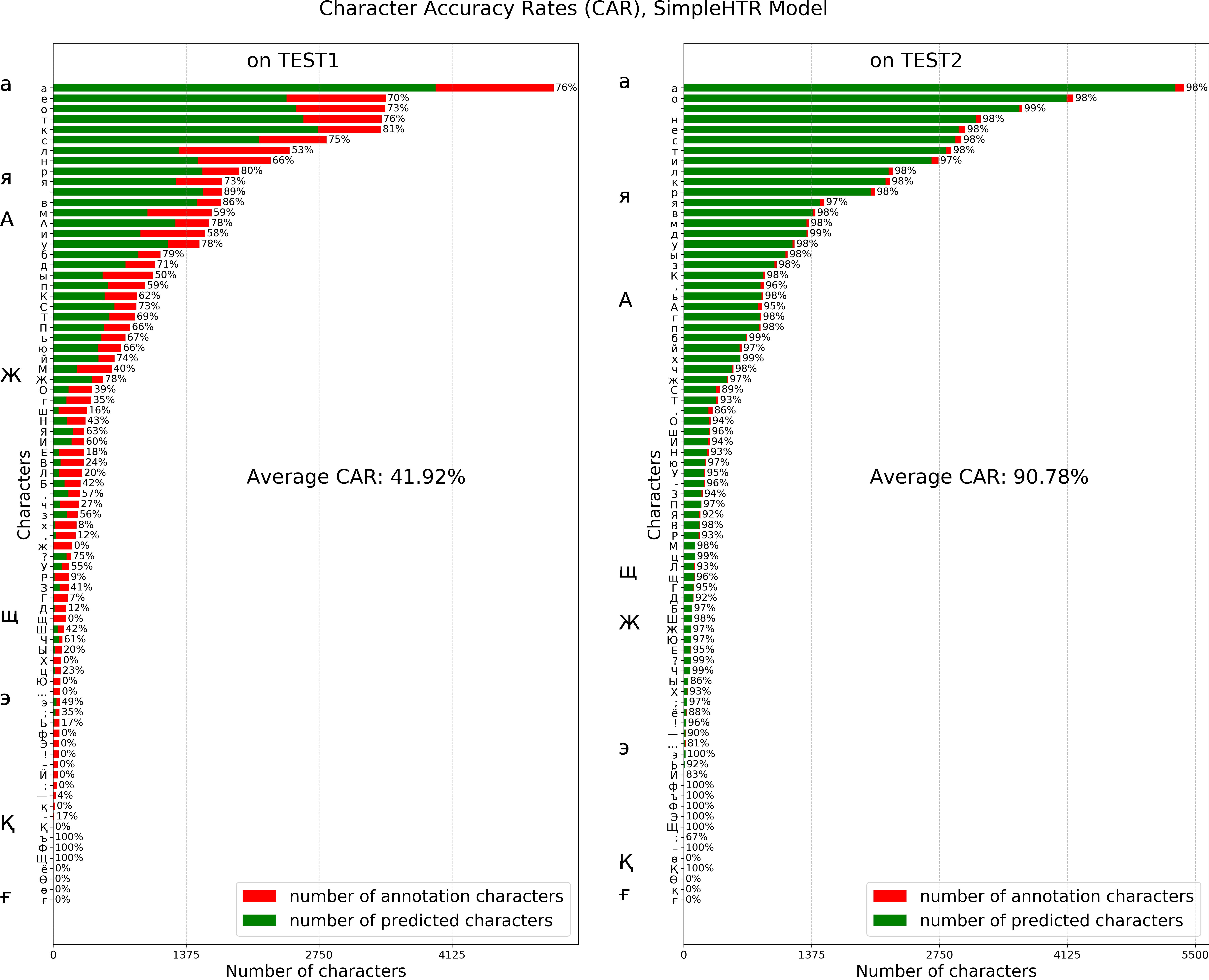}
	\caption{Character Accuracy Rate for SimpleHTR model } 
		\label{fig:car2_simplehtr}
	\end{figure}

\begin{table}[H]
\centering
	\caption{\footnotesize{RECOGNITION ACCURACY OF HANDWRITTEN CITY NAMES WITH VARIOUS DECODING METHODS}}

\begin{tabular}{|c|c|c|c|c|}
    \hline
    \multirow{2}{*}{Algorithm}& \multicolumn{2}{|c|}{ alignment of cursive }& \multicolumn{2}{|c|}{no alignment}\\
        \cline{2-5}
             & CER       & WAR  & CER & WAR  \\
\hline
    
     bestpath & 19.13 & 52.55 & 17.97 & 57.11  \\
    \hline
    beamsearch & 18.99 & 53.33  & 17.73 & 58.33 \\
    \hline
    wordbeamsearch & 16.38 & 73.55  & 15.78 & 75.11 \\
    \hline

\end{tabular}

\label{tab:veris}
\end{table}
Figure \ref{fig:South-Kazakhstan} shows an image with the name of the region that was submitted to the entrance, and in Figure \ref{fig:SimpleHTR_output} we can see the recognized word “South Kazakhstan” (“Yuzhno-Kazakhstanskaya” in Russian) with a probability of 86 percent.
\paragraph{Second Dataset (HKR Dataset)}
The SimpleHTR model showed in the first test of the dataset 20.13\% Character error rate (CER) and second dataset 1.55\% CER. 
We also evaluated the SimpleHTR model by each Character accuracy rate (Figure \ref{fig:car2_simplehtr}). Word error rate (WER) was 58.97\% for TEST1and 11.09\% for TEST2. The result for TEST2 shows that the model can recognize words that exist in the Training dataset but have completely different handwriting styles. The TEST1 dataset shows that the result is not good when the model recognizes the words that do not exist in Training and Validation datasets.


\subsection{Bluche and Puigcerver models}
After training, validation, and test datasets were prepared, and models were trained, comparative evaluation experiments were conducted. The Bluche and Puigcerver model were trained on the second dataset (HKR dataset). We evaluated these models by the standard performance measures used for all results presented: CER and WER. For all models the minibatch of 32 size and Early Stopping after 20 epochs without improvement in validation loss value and lr=0.001 were set. For the best use of each model, within the 20 tolerance epochs, ReduceLRonPlateau schedule\cite{vinciarelli2001new} with a decay factor of 0.2 after 10 epochs without improvement in validation loss value was also used.

Table \ref{tab:Bluche_Puigcerver} shows the result of comparison between the two models. We can observe that the Puigcerver has a higher error rate compared with the Bluche because the Puigcerver model has many parameters (~9.6M) and overfitting on the dataset.

\begin{table}[H]
\centering
	\caption{\footnotesize{CER, WER for Bluche and Puigcerver}}

\begin{tabular}{|c|c|c|c|c|}
    \hline
    \multirow{2}{*}{Algorithm}& \multicolumn{2}{|c|}{TEST1}& \multicolumn{2}{|c|}{TEST2}\\
        \cline{2-5}
             & CER       & WER  & CER & WER  \\
\hline
    
     Bluche & 16.15\% & 59.64\% &10.15\% &37.49\%  \\
    \hline
    Puigcerver & 73.43\% & 96.89\%  & 54.75\% & 82.91\% \\
    \hline

\end{tabular}

\label{tab:Bluche_Puigcerver}
\end{table}

\begin{figure}[H]
	\centering
	\includegraphics[width=\linewidth]{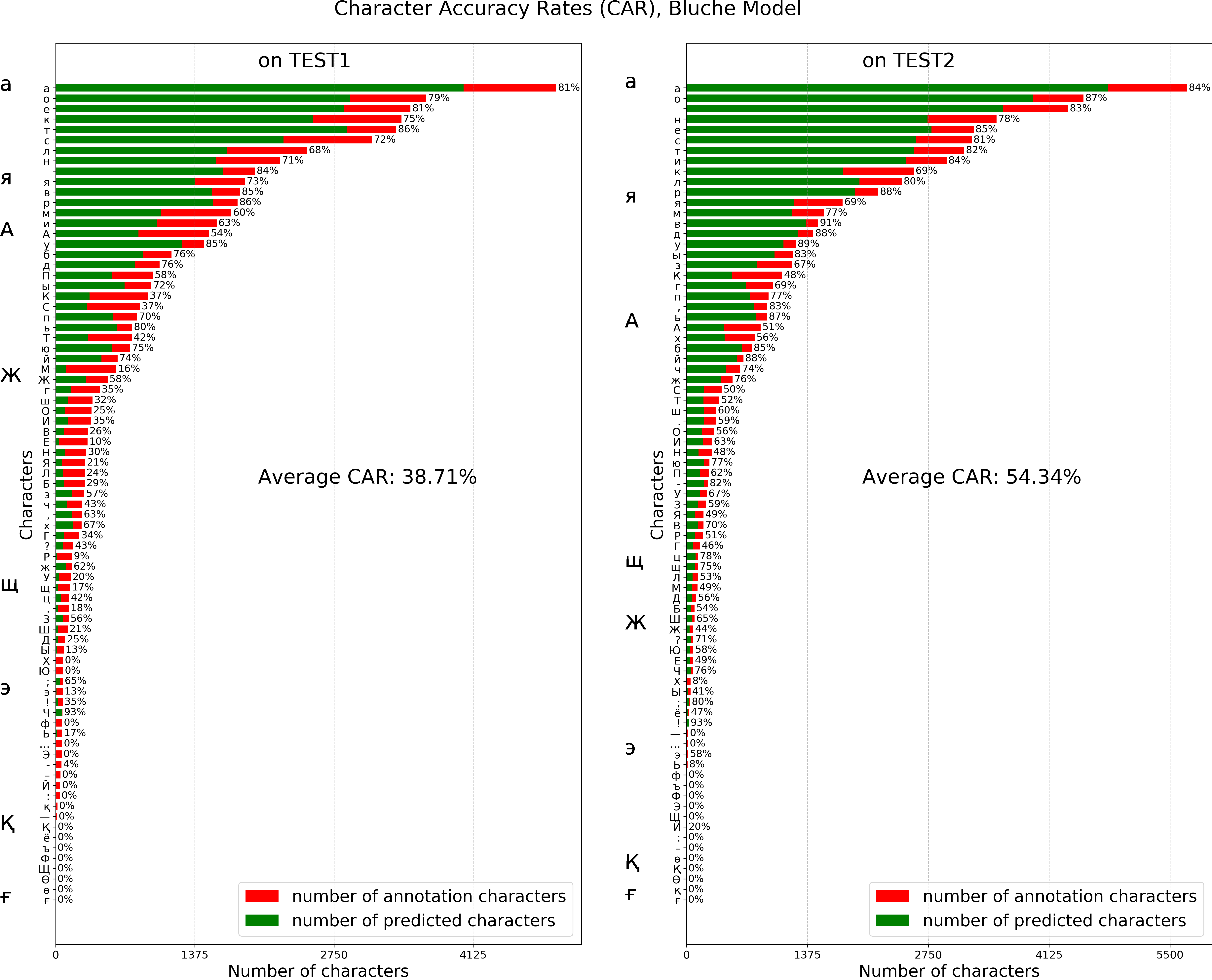}
	\caption{Bluche HTR model performance on TEST1 and TEST2 dataset} 
		\label{fig:Bluche_HTR}
\end{figure}

\begin{figure}[H]
	\centering
	\includegraphics[width=\linewidth]{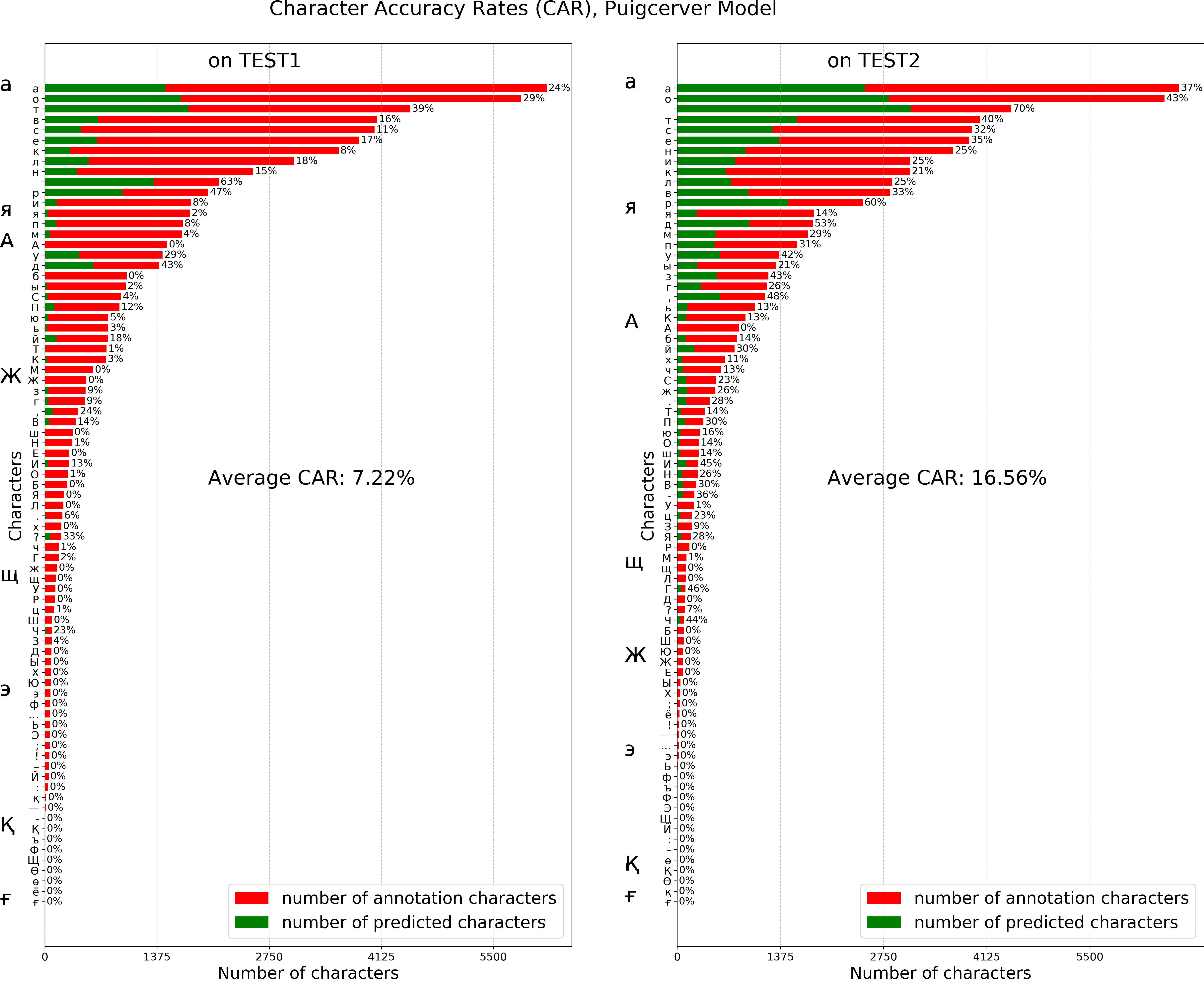}
	\caption{Puigcerver HTR model performance on TEST1 and TEST2 dataset } 
		\label{fig:Puigcerve_HTR}
\end{figure}

Figure  \ref{fig:Bluche_HTR} and \ref{fig:Puigcerve_HTR} present the character accuracy rate which shows how the model detects each character.


\section{Conclusion}
Two interrelated problems are considered in the paper: classification of handwritten names of cities and HTR using various deep learning models. The first model is used for classification of handwritten cities based on deep CNN and the other three models (SimpleHTR, Bluche, Puigcerver) are used for HTR which contains CNN layers, RNN layers, and the CTC decoding algorithm.

Experiments on classification of handwritten names of cities were conducted using various machine learning methods and the following results for recognition accuracy were obtained on the test  data: 1) 55.3\% for CNN; 2) 57.1\%  for SimpleHTR recurrent CNN using best-path decoding algorithms, 58.3\% for Beamsearch and 75.1\%  wordbeamsearch is The best result was shown by for Wordbeamsearch, which uses a dictionary for the final correction of the text under recognition.

Experiments on HTR were also conducted with various deep learning methods and the following results for recognition accuracy were obtained for the two test datasets: 1) Bluche model achieved 16.15\% CER and 59.64\% WER in the first test dataset and 10.15\% CER and 37.49\% WER in the second dataset; 2)The Puigcerver HTR model showed 73.43\% CER and 96.89\% WER in the first test dataset and 54.75\% CER and 82.91\% WER in the second dataset; 3)The SimpleHTR model showed 20.13\% CER and 58.97\% WER in the first test dataset and 1.55\% CER and 11.09\% WER in the second dataset.

\paragraph{Acknowledgment}
The work was carried out within the framework of the grant project No. AR05135175 with the support of the Ministry of Education and Science of the Republic of Kazakhstan.


\end{multicols}

\end{document}